\documentstyle[11pt,latex-acl]{article}
\author{David Yarowsky\thanks{
This research was supported by an NDSEG Fellowship, ARPA grant N00014-90-J-1863
and ARO grant DAAL 03-89-C0031 PRI.
The author is also affiliated with the Linguistics Research Department of
AT\&T Bell Laboratories, and greatly appreciates the use of its resources in
support of this work. He would like to thank Jason Eisner, Libby Levison, Mark
Liberman, Mitch Marcus, Joseph Rosenzweig and Mark Zeren
for their valuable feedback.
} \\
Department of Computer and Information Science \\
University of Pennsylvania \\
Philadelphia, PA 19104 \\
{\tt yarowsky@unagi.cis.upenn.edu}
}

\title{
\vspace{-0.6in}
     DECISION LISTS FOR LEXICAL AMBIGUITY \\
               RESOLUTION: \\
       Application to Accent Restoration \\
            in Spanish and French}

\begin{document}
\maketitle
\bibliographystyle{acl}

\begin{abstract}
This paper presents a statistical decision procedure
for lexical ambiguity resolution. The algorithm exploits
both local syntactic patterns and more distant collocational evidence,
generating an efficient, effective,
and highly perspicuous recipe for resolving a given ambiguity.
By identifying and utilizing only the single best disambiguating evidence
in a target context, the algorithm avoids the problematic complex modeling of
statistical dependencies.
Although directly applicable to a wide class of ambiguities, the
algorithm is described and evaluated in a realistic case study,
the problem of restoring missing accents in Spanish and French text.
Current accuracy exceeds 99\% on the full task, and typically is over 90\%
for even the most difficult ambiguities.
\end{abstract}

\section{INTRODUCTION}

This paper presents a general-purpose statistical decision procedure for
lexical ambiguity resolution based on decision lists (Rivest, 1987).
The algorithm considers multiple types of evidence in the context of
an ambiguous word, exploiting differences in collocational distribution
as measured by log-likelihoods.
Unlike standard Bayesian approaches, however,
it does not combine the log-likelihoods of all available pieces of contextual
evidence, but bases its classifications solely on the single most reliable
piece of evidence identified in the target context.
Perhaps surprisingly, this strategy appears to yield the same or
even slightly better precision than the combination of evidence approach
when trained on the same features. It also brings with it several additional
advantages, the greatest of which is the ability to include multiple,
highly non-independent sources of evidence without complex modeling of
dependencies. Some other advantages are significant simplicity and ease of
implementation, transparent understandability of the resulting decision list,
and easy adaptability to new
domains. The particular domain chosen here as a case study is the problem of
restoring missing accents\footnote{For brevity, the term {\it accent} will
typically refer to the general class of accents and other diacritics, including
\^{e},\`{e},\'{e},\"{o}}
to Spanish and French text.  Because it requires the
resolution of both semantic and syntactic ambiguity, and offers an objective
ground truth for automatic evaluation, it is particularly well suited for
demonstrating and testing the capabilities of the given algorithm. It is also
a practical problem with immediate application.

\section{PROBLEM DESCRIPTION}

The general problem considered here is the resolution of lexical ambiguity,
both syntactic and semantic,
based on properties of the surrounding context. Accent restoration is merely
an instance of a closely-related class of problems including word-sense
disambiguation, word choice selection in machine translation, homograph and
homophone disambiguation, and capitalization restoration. The given algorithm
may be used to solve each of these problems, and has been applied without
modification to the case of homograph disambiguation in speech
synthesis (Sproat, Hirschberg and Yarowsky, 1992).

It may not be immediately apparent to the reader why this set of problems
forms a natural class,
similar in origin and solvable by a single type of algorithm.
In each case it is necessary to
disambiguate two or more semantically distinct word-forms which
have been conflated into the same representation in some medium.

In the prototypical instance of this class, word-sense disambiguation,
such distinct semantic concepts as {\it river bank}, {\it financial bank}
and {\it to bank an airplane} are conflated in ordinary text.
Word associations and syntactic patterns are sufficient to identify and label
the correct form.  In homophone disambiguation,
distinct semantic concepts such as {\it ceiling} and {\it sealing} have also
become represented by the same ambiguous form, but in the medium of speech
and with similar disambiguating clues.

Capitalization restoration is a similar problem in that distinct
semantic concepts such as {\it AIDS}/{\it aids} (disease or helpful tools) and
{\it Bush}/{\it bush} (president or shrub) are ambiguous, but in the
medium of  all-capitalized (or casefree) text, which includes titles and the
beginning of sentences.
Note that what was once just a capitalization ambiguity between
{\it Prolog} (computer language) and {\it prolog} (introduction) has
is becoming a ``sense'' ambiguity since the computer language is now often
written in lower case, indicating the fundamental similarity of these
problems.

Accent restoration involves lexical ambiguity, such as between the
concepts {\it c\^{o}te} (coast) and  {\it c\^{o}t\'{e}} (side), in textual
mediums where accents are missing.  It is traditional in
Spanish and French for diacritics to be omitted from capitalized letters.
This is particularly a problem in all-capitalized text such as
headlines. Accents in on-line text may also be systematically
stripped by many computational processes which are not 8-bit clean (such
as some e-mail transmissions), and may be routinely omitted by Spanish and
French typists in informal computer correspondence.

Missing accents may create both semantic and syntactic ambiguities,
including tense or mood distinctions which may only
be resolved by distant temporal markers or non-syntactic cues.
The most common accent ambiguity in Spanish is between the endings
{\it -o} and \linebreak
{\it -\'{o}}, such as in the case of {completo} vs. {complet\'{o}}.
This is a present/preterite tense ambiguity for nearly all \linebreak
-ar verbs, and
very often also a part of speech ambiguity, as the -o form is a frequently a
noun as well.  The second most common general ambiguity is between the
past-subjunctive and future tenses of nearly all -ar verbs
(eg: {\it terminara} vs. {\it terminar\'{a}}),
both of which are 3rd person singular forms. This is a particularly challenging
class and is not readily amenable to traditional part-of-speech tagging
algorithms such as local trigram-based taggers.
Some purely semantic ambiguities include the nouns {\it secretaria} (secretary)
vs.  {\it secretar\'{\i}a} (secretariat),
{\it sabana} (grassland) vs. {\it s\'{a}bana} (bed sheet), and
{\it politica} (female politician) vs.  {\it pol\'{\i}tica} (politics).
The distribution of ambiguity types in French is similar. The most common case
is between {\it -e} and {\it -\'{e}}, which is both a past participle/present
tense ambiguity, and often a part-of-speech ambiguity (with nouns and
adjectives) as well.  Purely semantic ambiguities are more common
than in Spanish, and include {\it trait\'{e}}/{\it traite} (treaty/draft),
{\it marche}/{\it march\'{e}} (step/market), and the {\it cote} example
mentioned above.

Accent restoration provides several advantages as a case study for the
explication and evaluation of the proposed decision-list algorithm.
First, as noted above, it offers a broad spectrum of ambiguity types,
both syntactic and semantic, and shows the ability of the algorithm
to handle these diverse problems. Second, the correct accent pattern
is directly recoverable: unlimited
quantities of test material may be constructed by stripping the accents
from correctly-accented text and then using the original as a fully objective
standard for automatic evaluation. By contrast, in traditional word-sense
disambiguation, hand-labeling training and test data is a laborious and
subjective task.  Third, the task of restoring missing accents and resolving
ambiguous forms shows considerable commercial applicability,
both as a stand-alone application or part of the front-end to
NLP systems. There is also a large potential commercial market in its use
in grammar and spelling correctors, and in aids for inserting the proper
diacritics automatically when one types\footnote{Such a tool would be
particularly useful for typing Spanish or French on Anglo-centric computer
keyboards, where entering accents and other diacritic marks every few
keystrokes can be laborious.}.  Thus while accent restoration may not be be
the prototypical member of the class of lexical-ambiguity resolution problems,
it is an especially useful one for describing and evaluating a proposed
solution to this class of problems.

\section{PREVIOUS WORK}

The problem of accent restoration in text has received minimal coverage in the
literature, especially in English, despite its many interesting aspects.
Most work in this area appears to done in the form of in-house or commercial
software, so for the most part the problem and its potential solutions are
without comprehensive published analysis. The best treatment I've discovered is
from Fernand Marty (1986, 1992), who for more than a decade has been
painstakingly crafting a system which includes accent restoration as part of a
comprehensive system of syntactic, morphological and phonetic analysis,
with an intended application in French text-to-speech synthesis.
He incorporates information extracted from several French dictionaries and uses
basic collocational and syntactic evidence in hand-built rules and heuristics.
While the scope and complexity of this effort is remarkable,
this paper will focus on a solution to the problem which requires
considerably less effort to implement.

The scope of work in lexical ambiguity resolution is very large.
Thus in the interest of space, discussion will
focus on the direct historic precursors and sources of inspiration
for the approach presented here. The central
tradition from which it emerges is that of the Bayesian classifier (Mosteller
and Wallace, 1964). This was expanded upon by (Gale et al., 1992), and in a
class-based variant by (Yarowsky, 1992). Decision trees (Brown, 1991)
have been usefully applied to word-sense ambiguities, and HMM part-of-speech
taggers (Jelinek 1985, Church 1988, Merialdo 1990) have addressed the
syntactic ambiguities presented here.  Hearst (1991) presented an effective
approach to modeling local contextual evidence, while Resnik (1993) gave a
classic treatment of the use of word classes in selectional constraints.
An algorithm for combining syntactic and semantic evidence in lexical
ambiguity resolution has been realized in (Chang et al., 1992).
A particularly successful algorithm for integrating a wide diversity of
evidence types using error driven learning was presented in Brill (1993).
While it has been applied primarily to syntactic problems, it shows tremendous
promise for equally impressive results in the area of semantic ambiguity
resolution.

The formal model of decision lists was presented in (Rivest, 1987).
I have restricted feature conjuncts to a much narrower complexity
than allowed in the original model -- namely to word and class trigrams.
The current approach was initially presented in
(Sproat et al., 1992), applied to the problem of homograph resolution
in text-to-speech synthesis. The algorithm achieved 97\% mean accuracy
on a disambiguation task involving a sample of 13 homographs\footnote{
Baseline accuracy for this data (using the most common pronunciation) is
67\%.}.

\section{ALGORITHM}

\subsection{Step 1: Identify the Ambiguities in Accent Pattern}

Most words in Spanish and French exhibit only one accent pattern.
Basic corpus analysis will indicate which is the most common
pattern for each word, and may be used in conjunction with or
independent of dictionaries and other lexical resources.

The initial step is to take a histogram of a corpus {\it with} accents
and diacritics retained, and compute a table of accent pattern distributions
as follows:
\begin{center}
\begin{tabular}{|l|l|r|r|}
\hline
De-accented Form & Accent Pattern & \% & Number \\
\hline
cesse & cesse        & 53\% & 669 \\
      & cess\'{e}   & 47\% & 593 \\
\hline
cout  & co\^{u}t    & 100\% & 330 \\
\hline
couta  & co\^{u}ta   & 100\% & 41 \\
\hline
coute  & co\^{u}t\'{e} & 53\% & 107 \\
       & co\^{u}te     & 47\% & 96 \\
\hline
cote & c\^{o}t\'{e} & 69\% & 2645 \\
     & c\^{o}te & 28\% &  1040 \\
     & cote     & 3\% &  99  \\
     & cot\'{e} & $<$1\% & 15 \\
\hline
cotiere & c\^{o}ti\`{e}re & 100\% & 296 \\
\hline
\end{tabular}
\end{center}
\vspace{.1in}

For words with multiple accent patterns, steps 2-5 are applied.

\subsection{Step 2: Collect Training Contexts}

For a particular case of accent ambiguity identified above,
collect ${\pm}k$ words of context around all occurrences in the
corpus, label the concordance line with the observed accent pattern,
and then strip the accents from the data.  This will yield a training
set such as the following:
\begin{center}
\setlength{\tabcolsep}{0.5mm}
\begin{tabular}{|l|rl|}
\hline
Pattern & & Context \\
\hline
\ (1) c\^{o}t\'{e} \ &   du laisser de \ & {\it cote} \  faute de temps  \\
\ (1) c\^{o}t\'{e} &  appeler l' autre \ & {\it cote}  \ de l' atlantique  \\
\ (1) c\^{o}t\'{e} &  passe de notre \ & {\it cote} \  de la frontiere  \\
\hline
\ (2) c\^{o}te     &  vivre sur notre \ & {\it cote} \ ouest toujours verte \\
\ (2) c\^{o}te     &  creer sur la  \ & {\it cote}  \ du labrador des  \\
\ (2) c\^{o}te     &  travaillaient cote a \ & {\it cote}  \ , ils avaient  \\
\hline
\end{tabular}
\end{center}
\vspace{.1in}

The training corpora used in this experiment were the Spanish AP Newswire
(1991-1993, 49 million words), the French Canadian Hansards (1986-1988, 19
million words), and a collection from {\it Le Monde} (1 million words).

\subsection{Step \nolinebreak 3: \nolinebreak Measure \nolinebreak
Collocational \nolinebreak Distributions}

The driving force behind this disambiguation algorithm is the uneven
distribution of collocations\footnote{The term {\it collocation} is used here
in its broad sense, meaning words appearing adjacent to or near each other
(literally, in the same location), and does not imply only idiomatic or
non-compositional associations.} with respect to the ambiguous token being
classified.  Certain collocations will indicate one accent pattern, while
different collocations will tend to indicate another.
The goal of this stage of the algorithm is to measure a large number of
collocational distributions and select those which are most useful in
identifying the accent pattern of the ambiguous word.

The following are the initial types of collocations considered:

\begin{itemize}
\item Word immediately to the right (+1 W)
\item Word immediately to the left (-1 W)
\item Word found in ${\pm}k$ word window\footnote{
The optimal value of {\it k} is sensitive to the type of ambiguity.
Semantic or topic-based ambiguities warrant a larger window ($k \approx
20-50$), while more local syntactic ambiguities warrant a smaller window
($k \approx 3$ or $4$)} (${\pm}$k W)
\item Pair of words at offsets -2 and -1
\item Pair of words at offsets -1 and +1
\item Pair of words at offsets +1 and +2
\end{itemize}

For the two major accent patterns of the French word {\it cote},
below is a small sample of these distributions for several
types of collocations:

\begin{center}
\begin{tabular}{|l|l|c|c|}
\hline
Position & Collocation & {\bf c\^{o}te}  &  {\bf c\^{o}t\'{e}} \\
\hline
-1 {\sc w} & du {\it cote} & 0 &  536 \\
           & la {\it cote} & 766 & 1 \\
           & un {\it cote} & 0 & 216 \\
           & notre {\it cote} & 10 & 70 \\
\hline
+1 {\sc w} & {\it cote} ouest & 288 & 1 \\
           & {\it cote} est & 174 & 3 \\
           & {\it cote} du & 55 & 156 \\
\hline
+1{\sc w},+2{\sc w} & {\it cote} du gouvernement & 0 & 62 \\
-2{\sc w},-1{\sc w} & cote a {\it cote} & 23 & 0 \\
\hline
${\pm}$k {\sc w} & poisson (in ${\pm}k$ words)  & 20 & 0 \\
${\pm}$k {\sc w} & ports (in ${\pm}k$ words)  & 22 & 0 \\
${\pm}$k {\sc w} & opposition (in ${\pm}k$ words) & 0 & 39 \\
\hline
\end{tabular}
\end{center}
\vspace{.1in}

This core set of evidence presupposes no language-specific
knowledge. However, if additional language resources are available,
it may be desirable to include a larger feature set. For example,
if lemmatization procedures are available, collocational measures
for morphological roots will tend to yield more succinct and generalizable
evidence than measuring the distributions for each of the inflected forms.
If part-of-speech information is available in a lexicon, it is useful to
compute the distributions for part-of-speech bigrams and trigrams as above.
Note that it's not necessary to determine the actual parts-of-speech of
words in context; using only the most likely part of speech or a set
of all possibilities will produce adequate, if somewhat diluted, distributional
evidence. Similarly, it is useful to compute collocational statistics
for arbitrary word classes, such as the class {\sc weekday} =\{ {\it domingo},
{\it lunes}, {\it martes}, ... \}. Such classes may cover many types of
associations, and need not be mutually exclusive.

For the French experiments, no additional linguistic knowledge or lexical
resources were used.  The decision lists were trained solely on raw word
associations without additional patterns based on part of speech,
morphological analysis or word class. Hence the reported performance is
representative of what may be achieved with a rapid, inexpensive
implementation based strictly on the distributional properties of raw text.

For the Spanish experiments, a richer set of evidence was utilized. Use of a
morphological analyzer (developed by Tzoukermann and Liberman (1990)) allowed
distributional measures to be computed for associations of lemmas
(morphological roots), improving generalization to different inflected forms
not observed in the training data.  Also, a basic lexicon with possible
parts of speech (augmented by the morphological analyzer) allowed adjacent
part-of-speech sequences to be used as disambiguating evidence. A relatively
coarse level of analysis (e.g. {\sc noun}, {\sc adjective}, {\sc
subject-pronoun}, {\sc article}, etc.), augmented with independently
modeled features representing gender, person, and number, was found to be
most effective. However, when a word was listed with multiple parts-of-speech,
no relative frequency distribution was available.  Such words were given a
part-of-speech tag consisting of the union of the possibilities (eg {\sc
adjective-noun}), as in Kupiec (1989).  Thus sequences of pure part-of-speech
tags were highly reliable, while the potential sources of noise were isolated
and modeled separately.  In addition, several word classes such as {\sc
weekday} and {\sc month} were defined, primarily focusing on time words
because so many accent ambiguities involve tense distinctions.

To build a full part of speech tagger for Spanish would be quite costly
(and require special tagged corpora). The current approach uses just the
information available in dictionaries, exploiting only that which is useful
for the accent restoration task.  Were dictionaries not available, a
productive approximation could have been made using the associational
distributions of suffixes (such as {\it -aba}, {\it -aste}, {\it -amos})
which are often satisfactory indicators of part of speech in morphologically
rich languages such as Spanish.

The use of the word-class and part-of-speech data is illustrated below,
with the example of distinguishing {\it terminara}/{\it terminar\'{a}}
(a subjunctive/future tense ambiguity):

\begin{center}
\begin{tabular}{|l|c|c|}
\hline
Collocation & {\bf termin-}  &  {\bf termin-} \\
            & {\bf ara}  &  {\bf ar\'{a}} \\
\hline
{\sc preposition que} {\it terminara} & 31 & 0 \\
de que {\it terminara} & 15 & 0 \\
para que {\it terminara} & 14 & 0 \\
\hline
{\sc noun que} {\it terminara} & 0 & 13 \\
carrera que {\it terminara} & 0 & 3 \\
reunion que {\it terminara} & 0 & 2 \\
acuerdo que {\it terminara} & 0 & 2 \\
\hline
que {\it terminara} & 42 & 37 \\
\hline
{\sc weekday} (within ${\pm}k$ words) & 0 & 23 \\
domingo (within ${\pm}k$ words) & 0 & 10 \\
viernes (within ${\pm}k$ words) & 0 & 4 \\
\hline
\end{tabular}
\end{center}

\subsection{Step 4: Sort by Log-Likelihood into Decision Lists}

The next step is to compute the ratio called the {\it log-likelihood}:

\[ Abs ( Log ( \frac{Pr(Accent\_Pattern_1|Collocation_i)}
                    {Pr(Accent\_Pattern_2|Collocation_i)} ) ) \]

The collocations most strongly indicative of a particular pattern will have
the largest log-likelihood. Sorting by this value will list the strongest
and most reliable evidence first\footnote{
Problems arise when an observed count is 0. Clearly the probability of seeing
{\it c\^{o}t\'{e}} in the context of {\it poisson} is not 0, even though
no such collocation was observed in the training data. Finding a more accurate
probability estimate depends on several factors, including the size of the
training sample, nature of the collocation (adjacent bigrams or wider context),
our prior expectation about the similarity of contexts, and the amount of
noise in the training data.  Several smoothing methods have been explored here,
including those discussed in (Gale et al., 1992).
In one technique, all observed distributions with the same 0-denominator raw
frequency ratio (such as 2/0) are taken collectively, the average agreement
rate of these distributions with additional held-out training data is measured,
and from this a more realistic
estimate of the likelihood ratio (e.g. 1.8/0.2) is computed.
However, in the simplest implementation, satisfactory results may be achieved
by adding a small constant $\alpha$ to the numerator and denominator,
where $\alpha$ is selected empirically to optimize classification performance.
For this data, relatively small $\alpha$ (between 0.1 and 0.25) tended to be
effective, while noisier training data warrant larger $\alpha$.}.

Evidence sorted in the above manner will yield a decision list like
the following, highly abbreviated example\footnote{Entries marked with \dag
\ are pruned in Step 5, below.}:

\begin{center}
\setlength{\tabcolsep}{0.5mm}
\begin{tabular}{|r|ll|}
\hline
\ LogL      \ & \ Evidence &  Classification \ \\
\hline
8.28 \ & \ {\sc preposition que} {\it terminara} & $\Rightarrow$ terminara \ \\
\dag 7.24 \ & \ de que {\it terminara} & $\Rightarrow$ terminara \ \\
\dag 7.14 \ & \ para que {\it terminara} & $\Rightarrow$ terminara \ \\
6.87 \ & \ y {\it terminara} & $\Rightarrow$ terminar\'{a} \ \\
6.64 \ & \ {\sc weekday} (within ${\pm}k$ words) & $\Rightarrow$ terminar\'{a}
\ \\
5.82 \ & \ {\sc noun que} {\it terminara} & $\Rightarrow$ terminar\'{a} \ \\
\dag 5.45 \ & \ domingo (within ${\pm}k$ words) & $\Rightarrow$ terminar\'{a} \
\\
\hline
\end{tabular}
\end{center}

The resulting decision list is used to classify new
examples by identifying the highest line in the list that
matches the given context and returning the indicated classification.
See Step 7 for a full description of this process.

\subsection{Step 5: Optional Pruning and Interpolation}

A potentially useful optional procedure is the interpolation
of log-likelihood ratios between those computed from the full
data set (the {\it global} probabilities) and those computed from
the residual training data left at a given point in the decision list
when all higher-ranked patterns failed to match (i.e. the {\it residual}
probabilities).  The residual probabilities are more relevant, but
since the size of the residual training data shrinks at each level
in the list, they are often much more poorly estimated (and in many
cases there may be no relevant data left in the residual on which to
compute the distribution of accent patterns for a given collocation).
In contrast,  the global probabilities are better estimated but less relevant.
A reasonable compromise is to interpolate between the two, where the
interpolated estimate is $\beta \times global + \gamma \times residual$.
When the residual probabilities are based on a large training set and
are well estimated, $\gamma$ should dominate, while in cases the relevant
residual is small or non-existent, $\beta$ should dominate.  If always
$\beta=0$ and $\gamma=1$ (exclusive use of the residual), the result
is a degenerate (strictly right-branching) decision tree with
severe sparse data problems.  Alternately, if one assumes that
likelihood ratios for a given collocation are functionally equivalent
at each line of a decision list, then one could exclusively use the
global (always $\beta=1$ and $\gamma=0$). This is clearly the easiest
and fastest approach, as probability distributions do not need to
be recomputed as the list is constructed.  Which approach
is best? Using only the global proabilities does surprisingly well,
and the results cited here are based on this readily replicatable procedure.
The reason is grounded in the strong tendency of a word to exhibit only
one sense or accent pattern per collocation (discussed in Step 7 and
(Yarowsky, 1993)). Most classifications are based on a $x$ vs. 0 distribution,
and while the magnitude of the log-likelihood ratios may decrease in the
residual, they rarely change sign. There are cases where this
does happen and it appears that some interpolation helps, but for {\it this}
problem the relatively small difference in performance does not seem to
justify the greatly increased computational cost.

Two kinds of optional pruning can also increase
the efficiency of the decision lists.
The first handles the problem of ``redundancy by subsumption,'' which is
clearly visible in the example decision lists above (in {\sc weekday} and
{\it domingo}). When lemmas and word-classes precede their member words in
the list, the latter will be ignored and can be pruned. If a bigram is
unambiguous, probability distributions for dependent trigrams will not
even be generated, since they will provide no additional information.

The second, pruning in a cross-validation phase, compensates for the
minimal observed over-modeling of the data. Once a decision list is built
it is applied to its own training set plus some held-out cross-validation
data ({\it not} the test data). Lines in the list which contribute to more
incorrect classifications than correct ones are removed. This also indirectly
handles problems that may result from the omission of the interpolation step.
If space is at a premium, lines which are never used in the cross-validation
step may also be pruned.  However, useful information is lost here, and words
pruned in this way may have contributed to the classification of testing
examples. A 3\% drop in performance is observed, but an over 90\% reduction
in space is realized. The optimum pruning strategy is subject to cost-benefit
analysis.  In the results reported below, all pruning except this final
space-saving step was utilized.

\subsection{Step 6: Train Decision Lists for General Classes of Ambiguity}

For many similar types of ambiguities, such
as the Spanish subjunctive/future distinction between {\it -ara} and {\it
ar\'{a}}, the decision lists for individual cases will be quite similar and
use the same basic evidence for the classification (such as presence of
nearby time adverbials).  It is useful to build a general decision list for
all {\it -ara}/{\it ar\'{a}} ambiguities.  This also tends to improve
performance on words for which there is inadequate training data to build a
full individual decision lists. The process for building this general
class disambiguator is basically identical to that described in Steps 2-5
above, except that in Step 2, training contexts are pooled for all individual
instances of the class (such as all {\it -ara}/{\it -ar\'{a}} ambiguities).  It
is important to give each individual {\it -ara} word roughly equal
representation in the training set, however, lest the list model the
idiosyncrasies of the most frequent class members, rather than identify
the shared common features representative of the full class.

In Spanish, decision lists are trained for the general ambiguity classes
including {\it -o}/{\it -\'{o}}, {\it -e}/{\it -\'{e}},
{\it -ara}/{\it -ar\'{a}}, and {\it -aran}/{\it -ar\'{a}n}.  For each
ambiguous word belonginging to one of these classes, the accuracy of the
word-specific decision list is compared with the class-based list. If the
class's list performs adequately it is used. Words with idiosyncrasies that
are not modeled well by the class's list retain their own word-specific
decision list.

\subsection{Step 7: Using the Decision Lists}

Once these decision lists have been created, they may be used in
real time to determine the accent pattern for ambiguous words
in new contexts.

At run time, each word encountered in a text is looked up in a table.
If the accent pattern is unambiguous, as determined in Step 1,
the correct pattern is printed.
Ambiguous words have a table of the possible accent patterns and a pointer
to a decision list, either for that specific word or its ambiguity class
(as determined in Step 6).
This given list is searched for the highest ranking match in the word's
context, and a classification number is returned, indicating the most likely
of the word's accent patterns given the context\footnote{If all entries in
a decision list fail to match in a particular new context, a final entry
called {\sc default} is used; it indicates the most likely accent pattern
in cases where nothing matches.}.

{}From a statistical perspective, the evidence at the top of this list
will most reliably disambiguate the target word.  Given a word in a new
context to be assigned an accent pattern, if we may only base the
classification on a single line in the decision list, it should be the highest
ranking pattern that is present in the target context.
This is uncontroversial, and is solidly based in Bayesian decision
theory.

The question, however, is what to do with the less-reliable evidence that may
also be present in the target context. The common tradition is to combine the
available evidence in a weighted sum or product. This is done by Bayesian
classifiers, neural nets, IR-based classifiers and N-gram part-of-speech
taggers.  The system reported here is unusual in that it does no such
combination.  {\it Only} the single most reliable piece of evidence matched
in the target context is used. For example, in a context
of {\it cote} containing {\it poisson}, {\it ports} and {\it atlantique}, if
the adjacent feminine article {\it la cote} (the coast) is present, only this
best evidence is used and the supporting semantic information ignored. Note
that if the masculine article {\it le cote} (the side) were present in a
similar maritime context, the most reliable evidence (gender agreement) would
override the semantic clues which would otherwise dominate if all evidence was
combined.  If no gender agreement constraint were present in that context, the
first matching semantic evidence would be used.

There are several motivations for this approach. The first is that combining
all available evidence rarely produces a different classification than just
using the single most reliable evidence, and when these differ it is as likely
to hurt as to help. In a study comparing results for 20 words in a binary
homograph disambiguation task, based strictly on words in local ($\pm 4$ word)
context, the following differences were observed between an algorithm taking
the single best evidence, and an otherwise identical algorithm combining all
available matching evidence:\footnote{In cases of disagreement, using the
single best evidence outperforms the combination of evidence 65\% to 35\%.
This observed difference is 1.9 standard deviations greater than
expected by chance and is statistically significant.
}

\begin{center}
\begin{tabular}{|rl|r|}
\multicolumn{3}{l}{\bf Combining vs. Not Combining Probabilities} \\
\hline
Agree - & Both classifications correct & 92\% \\
      & Both classifications incorrect & 6\% \\
\hline
Disagree - & Single best evidence correct & 1.3\% \\
         & Combined evidence correct    & 0.7\% \\
\hline
Total -   &                              & 100\% \\
\hline
\end{tabular}
\end{center}

Of course that this behavior does not hold for all classification tasks, but
{\it does} seem to be characteristic of lexically-based word classifications.
This may be explained by the empirical observation that in most cases, and with
high probability, words exhibit only one {\it sense} in a given collocation
(Yarowsky, 1993).  Thus for this type of ambiguity resolution,
there is no apparent detriment, and some apparent performance gain, from using
only the single most reliable evidence in a classification. There are
other advantages as well, including run-time efficiency and ease of
parallelization.  However, the greatest gain comes from the ability to
incorporate multiple, non-independent information types in the decision
procedure. As noted above, a given word in context (such as {\it Castillos})
may match several times in the decision list, once for its parts of speech,
lemma, capitalized and capitalization-free forms, and possible word-classes
as well. By only using one of these matches, the gross exaggeration of
probability from combining all of these non-independent log-likelihoods is
avoided. While these dependencies may be modeled and corrected for in
Bayesian formalisms,  	it is difficult and costly to do so. Using only one
log-likelihood ratio without combination frees the algorithm to include a
wide spectrum of highly non-independent information without additional
algorithmic complexity or performance loss.

\section{EVALUATION}

Because we have only stripped accents artificially for testing purposes,
and the ``correct'' patterns exist on-line in the original corpus,
we can evaluate performance objectively and automatically.
This contrasts with other classification tasks such
as word-sense disambiguation and part-of-speech tagging, where at some point
human judgements are required. Regrettably, however, there are
errors in the original corpus, which can be quite substantial depending on
the type of accent. For example, in the Spanish data,
accents over the i (\'{\i}) are frequently
omitted; in a sample test 3.7\% of the appropriate \'{\i} accents were missing.
Thus the following results must be interpreted as agreement rates with
the corpus accent pattern; the true percent correct may be several
percentage points higher.

The following table gives a breakdown of the different types of Spanish
accent ambiguities, their relative frequency in the training corpus, and
the algorithm's performance on each:\footnote{
The term {\it prior} is a measure of the baseline performance one would
expect if the algorithm always chose the most common option.}

\begin{center}
\begin{tabular}{|l|r|r|r|}
\multicolumn{4}{l}{\bf Summary of Performance on Spanish:} \\
\hline
\multicolumn{4}{|l|}{Ambiguous Cases (18\% of tokens):} \\
\hline
Type & Freq. & Agreement & Prior \\
\hline
{\it -o}/{\it -\'{o}} & 81 \% & 98 \% & 86\% \\
{\it -ara}/{\it -ar\'{a}},{\it -aran}/{\it -ar\'{an}} & 4 \% & 92 \% & 84\% \\
Function Words & 13 \% & 98 \% & 94\% \\
Other & 2 \% & 97 \% & 95\% \\
\hline
Total &   & 98 \% & 93\% \\
\hline
\hline
\multicolumn{4}{|l|}{Unambiguous Cases (82\% of tokens):} \\
\hline
         &   & 100 \% & 100\% \\
\hline
\hline
Overall Performance: &   & 99.6 \% & 98.7\% \\
\hline
\end{tabular}
\end{center}

As observed before, the prior probabilities in favor of the most common
accent pattern are highly skewed, so one does reasonably well at this
task by always using the most common pattern. But the error rate is
still roughly 1 per every 75 words, which is unacceptably high. This
algorithm reduces that error rate by over 65\%. However, to get a
better picture of the algorithm's performance, the following
table gives a breakdown of results for a random set of the most problematic
cases -- words exhibiting the largest absolute number of the non-majority
accent patterns.  Collectively they constitute the most common potential
sources of error.

\begin{center}
\begin{tabular}{|l|l|r|c|r|}
\multicolumn{5}{c}{\bf Performance on Individual Ambiguities} \\
\multicolumn{5}{l}{\bf Spanish:} \\
\hline
Pattern 1 & Pattern 2 & Agrmnt & Prior & N \\
\hline
anuncio & anunci\'{o}      & 98.4\% & 57\%  & 9459 \\
registro & registr\'{o}   & 98.4\% & 60\%  & 2596 \\
marco & marc\'{o}         & 98.2\% & 52\%  & 2069 \\
completo & complet\'{o}    & 98.1\% & 54\%  & 1701 \\
retiro & retir\'{o}       & 97.5\% & 56\%  & 3713 \\
duro & dur\'{o}            & 96.8\% & 52\%  & 1466 \\
paso & pas\'{o}            & 96.4\% & 50\%  & 6383 \\
regalo & regal\'{o}        & 90.7\% & 56\%  &  280 \\
\hline
terminara & terminar\'{a}  & 82.9\% & 59\%  &  218 \\
llegara & llegar\'{a}      & 78.4\% & 64\%  &  860 \\
\hline
deje & dej\'{e}            & 89.1\% & 68\%  &  313 \\
gane & gan\'{e}            & 80.7\% & 60\%  &  279 \\
\hline
secretaria & secretar\'{\i}a & 84.5\% & 52\%  &  1065 \\
\hline
seria & ser\'{\i}a           & 97.7\% & 93\%  &  1065 \\
hacia & hac\'{\i}a           & 97.3\% & 91\%  &  2483 \\
esta & est\'{a}              & 97.1\% & 61\%  &  14140 \\
mi & m\'{\i}                 & 93.7\% & 82\%  &  1221 \\
\hline
\multicolumn{5}{l}{\bf French:} \\
\hline
cess\'{e} & cesse            & 97.7\% & 53\%  &  1262 \\
d\'{e}cid\'{e} & d\'{e}cide  & 96.5\% & 64\%  &  3667 \\
laisse & laiss\'{e}          & 95.5\% & 50\%  &  2624 \\
commence & commenc\'{e}      & 95.2\% & 54\%  &  2105 \\
\hline
c\^{o}t\'{e} & c\^{o}te      & 98.1\% & 69\%  &  3893 \\
trait\'{e} &   traite        & 95.6\% & 71\%  &  2865 \\
\hline
\end{tabular}
\end{center}

Evaluation is based on the corpora described in the algorithm's Step 2.
In all experiments, 4/5 of the data was used for training and the remaining
1/5 held out for testing. More accurate measures of algorithm performance were
obtained by repeating each experiment  5 times, using a different 1/5 of the
data for each test, and averaging the results. Note that in every experiment,
results were measured on independent test data not seen in the training phase.

It should be emphasized that the actual percent correct is
higher than these agreement figures, due to errors in the
original corpus. The relatively low agreement rate on words
with accented i's (\'{\i}) is a result of this.
To study this discrepancy further, a human judge fluent in Spanish
determined whether the corpus or decision list algorithm was
correct in two cases of disagreement. For the ambiguity case of
{\it mi}/{\it m\'{\i}}, the corpus was incorrect in 46\% of the disputed
tokens.  For the ambiguity {\it anuncio}/{\it anunci\'{o}},
the corpus was incorrect in 56\% of the disputed tokens.
I hope to obtain a more
reliable source of test material. However, it does appear that
in some cases the system's precision may rival that of the AP Newswire's
Spanish writers and translators.

\section{DISCUSSION}

The algorithm presented here has several advantages which make it
suitable for general lexical disambiguation tasks that require integrating
both semantic and syntactic distinctions. The incorporation
of word (and optionally part-of-speech) trigrams allows the modeling of many
local syntactic constraints, while collocational evidence in a wider context
allows for more semantic distinctions.  A key advantage of this approach is
that it allows the use of  multiple, highly  non-independent evidence types
(such as root form, inflected form, part of speech, thesaurus category or
application-specific clusters) and does so in a way that
avoids the complex modeling of statistical dependencies.  This allows the
decision lists to find the level of representation that best matches the
observed probability distributions. It is a kitchen-sink approach of the best
kind -- throw in many types of potentially relevant features and watch what
floats to the top.  While there are certainly other ways to combine such
evidence, this approach has many advantages.  In particular, precision seems
to be at least as good as that achieved with Bayesian methods applied to the
same evidence. This is not surprising, given the observation in (Leacock et
al., 1993) that widely divergent sense-disambiguation algorithms tend to
perform roughly the same given the same evidence.  The distinguishing criteria
therefore become:

\begin{itemize}
\item How readily can new and multiple types of evidence be incorporated into
      the algorithm?
\item How easy is the output to understand?
\item Can the resulting decision procedure be easily edited by hand?
\item Is it simple to implement and replicate, and can it be applied quickly
      to new domains?
\end{itemize}

The current algorithm rates very highly on all these standards of evaluation,
especially relative to some of the impenetrable black boxes produced by many
machine learning algorithms.  Its output is highly perspicuous: the resulting
decision list is organized like a recipe, with the most useful evidence first
and in highly readable form.  The generated decision procedure is also easy to
augment by hand, changing or adding patterns to the list. The algorithm is also
extremely flexible -- it is quite straightforward to use any new feature for
which a probability distribution can be calculated. This is a considerable
strength relative to other algorithms which are more constrained in their
ability to handle diverse types of evidence.  In a comparative study
(Yarowsky, 1994), the decision list algorithm outperformed both an N-Gram
tagger and Bayesian classifier primarily because it could effectively
integrate a wider range of available evidence types than its competitors.
Although a part-of-speech tagger exploiting gender and number agreement
might resolve many accent ambiguities, such constraints will fail to apply
in many cases and are difficult to apply generally, given the the problem of
identifying agreement relationships.  It would also be at considerable cost,
as good taggers or parsers typically involve several person-years of
development, plus often expensive proprietary lexicons and hand-tagged
training corpora.  In contrast, the current algorithm could be applied
quite quickly and cheaply to this problem. It was originally developed for
homograph disambiguation in text-to-speech synthesis (Sproat et al., 1992),
and was applied to the problem of accent restoration with virtually no
modifications in the code. It was applied to a new language, French, in a
matter of days and with no special lexical resources or linguistic knowledge,
basing its performance upon a strictly self-organizing analysis of the
distributional properties of French text. The flexibility and generality
of the algorithm and its potential feature set makes it readily applicable
to other problems of recovering lost information from text corpora;
I am currently pursuing its application to such problems as capitalization
restoration and the task of recovering vowels in Hebrew text.

\section{CONCLUSION}

This paper has presented a general-purpose algorithm for lexical ambiguity
resolution that is perspicuous, easy to implement, flexible and applied
quickly to new domains.  It incorporates class-based models at several levels,
and while it requires no special lexical resources or linguistic knowledge,
it effectively and transparently incorporates those which are available.
It successfully integrates part-of-speech patterns with local and
longer-distance collocational information to resolve both semantic and
syntactic ambiguities.  Finally, although the case study of accent
restoration in Spanish and French was chosen for its diversity of ambiguity
types and plentiful source of data for fully automatic and objective
evaluation, the algorithm solves a worthwhile problem in its own right with
promising commercial potential.


\begin{thebibliography}{}

\bibitem{X} Brill, Eric, ``A Corpus-Based Approach to Language Learning,''
Ph.D. Thesis, University of Pennsylvania, 1993.

\bibitem{X} Brown, Peter, Stephen Della Pietra, Vincent Della Pietra,
and Robert Mercer, ``Word Sense Disambiguation using Statistical Methods,''
{\it Proceedings of the 29th Annual Meeting of the Association for
Computational Linguistics}, pp. 264-270, 1991.

\bibitem{X} Chang, Jing-Shin, Yin-Fen Luo and Keh-Yih Su, ``GPSM: A Generalized
Probabilistic Semantic Model for Ambiguity Resolution,'' in
{\it Proceedings of the 30th Annual Meeting of the Association for
Computational Linguistics}, pp. 177-184, 1992.

\bibitem{X} Church, K.W., ``A Stochastic Parts Program and Noun Phrase Parser
for Unrestricted Text,'' in {\it Proceedings of the Second Conference on
Applied Natural Language Processing}, {\it ACL}, 136-143, 1988.

\bibitem{X} Gale, W., K. Church, and D. Yarowsky,  ``A Method for
Disambiguating Word Senses in a Large Corpus,'' {\it Computers and the
Humanities}, 26, 415-439, 1992.

\bibitem{X} Hearst, Marti, ``Noun Homograph Disambiguation Using Local Context
in Large Text Corpora,'' in {\it Using Corpora}, University of Waterloo,
Waterloo, Ontario, 1991.

\bibitem{X} Jelinek, F., ``Markov Source Modeling of Text Generation,''
in {\it Impact of Processing Techniques on Communication},
J. Skwirzinski, ed., Dordrecht, 1985.

\bibitem{X} Kupiec, Julian,  ``Probabilistic Models of Short and Long Distance
Word Dependencies in Running Text,'' in
{\it Proceedings, DARPA Speech and Natural Language Workshop}, Philadelphia,
February, pp. 290-295, 1989.

\bibitem{X} Leacock, Claudia, Geoffrey Towell and Ellen Voorhees ``Corpus-Based
Statistical Sense Resolution,'' in {\it Proceedings, ARPA Human Language
Technology Workshop}, 1993.

\bibitem{X} Marty, Fernand, ``Trois syst\`{e}mes informatiques de transcription
phon\'{e}tique et graph\'{e}mique'', in {\it Le Fran\c{c}ais Moderne},
pp. 179-197, 1992.

\bibitem{X} Marty, F. and R.S. Hart, ``Computer Program to Transcribe French
Text into Speech: Problems and Suggested Solutions'', Technical Report No.
LLL-T-6-85. Language Learning Laboratory; University of
Illinois. Urbana, Illinois, 1985.

\bibitem{X} Merialdo, B., `Tagging Text with a Probabilistic Model,'' in
{\it Proceedings of the IBM Natural Language ITL}, Paris, France,  pp.
161-172, 1990.

\bibitem{X} Mosteller, Frederick, and David Wallace, {\it Inference and
Disputed Authorship: The Federalist}, Addison-Wesley, Reading, Massachusetts,
1964.

\bibitem{X} Resnik, Philip, ``Selection and Information: A Class-Based Approach
to Lexical Relationships,'' Ph.D. Thesis, University of Pennsylvania, 1993.

\bibitem{X} Rivest, R. L., ``Learning Decision Lists,'' in {\it Machine
Learning}, 2, 229-246, 1987.

\bibitem{X} Sproat, Richard, Julia Hirschberg and David Yarowsky,
``A Corpus-based Synthesizer,'' in {\it Proceedings, International
Conference on Spoken Language Processing}, Banff, Alberta, October 1992.

\bibitem{X} Tzoukermann, Evelyne and Mark Liberman,
`` A Finite-state Morphological Processor for Spanish,''
in {\it Proceedings, COLING-90}, Helsinki, 1990.

\bibitem{X} Yarowsky, David,  ``Word-Sense Disambiguation
Using Statistical Models of Roget's Categories Trained on Large Corpora,''
in {\it Proceedings, COLING-92}, Nantes, France, 1992.

\bibitem{X} Yarowsky, David,  ``One Sense Per Collocation,'' in
{\it Proceedings, ARPA Human Language Technology Workshop}, Princeton, 1993.

\bibitem{X} Yarowsky, David,  ``A Comparison of Corpus-based Techniques
for Restoring Accents in Spanish and French Text,'' to appear in
{\it Proceedings, 2nd Annual Workshop on Very Large Text Corpora}, Kyoto,
Japan, 1994.


\end{thebibliography}
\end{document}